%% file: root.tex
\documentclass[letterpaper, 10 pt, journal, twoside]{IEEEtran}
\IEEEoverridecommandlockouts                              

\usepackage{amsfonts} 
\usepackage{amsmath}
\usepackage{amssymb}
\usepackage{amsthm}
\usepackage[ruled, lined, linesnumbered, longend]{algorithm2e}
\usepackage{array,multirow}
\usepackage{graphicx}
\usepackage[caption=false,font=small,labelfont=sf,textfont=sf]{subfig}
\usepackage[font=small,labelfont=bf]{caption}
\usepackage{textcomp}
\usepackage{stfloats}
\usepackage{verbatim}
\usepackage{tabularx}
\usepackage{xcolor}
\usepackage{siunitx}
\usepackage{makecell}
\usepackage{comment}
\usepackage{float}
\usepackage{booktabs}
\usepackage{algpseudocode}

\SetCommentSty{mycommfont}
\usepackage{mathtools}
\usepackage{bm}
\usepackage{epsfig} 
\usepackage{times}
\usepackage{listings}
\usepackage{psfrag}
\usepackage{xparse}
\usepackage[english]{babel}
\usepackage{macros}
\usepackage{url}
\usepackage{cite}
\usepackage{color}
\usepackage{cancel}
\usepackage{dirtytalk}
\usepackage[flushleft]{threeparttable}
\usepackage[figuresright]{rotating}
\usepackage[colorlinks,linkcolor=black,citecolor=black,urlcolor=black,bookmarks=false,hypertexnames=true]{hyperref} 

\definecolor{Gray}{gray}{0.90}
\newcolumntype{g}{>{\columncolor{Gray}}c}
\definecolor{ffe1da}{RGB}{255,225,218}
\definecolor{F7E0D5}{RGB}{247,224,213}
\definecolor{darkF7E0D5}{RGB}{209,154,128}

\definecolor{Dark}{rgb}{0,0,0}

\definecolor{OutdoorDark}{rgb}{0,.5,0}
\definecolor{comment}{rgb}{0.6, 0.4, 0.8}
\definecolor{IndoorDark}{rgb}{0,0.3,0.8}
\definecolor{SubTDark}{rgb}{0.5,.27,0.11}
\definecolor{AerialDark}{rgb}{.5,.0,.5}
\definecolor{UnderWaterDark}{rgb}{0.16, 0.46, 0.81}
\colorlet{Outdoor}{OutdoorDark!20!white}
\colorlet{Indoor}{IndoorDark!20!white}
\colorlet{SubT}{SubTDark!20!white}
\colorlet{Aerial}{AerialDark!20!white}
\colorlet{UnderWater}{UnderWaterDark!20!white}
\colorlet{OutdoorLight}{OutdoorDark!70!white}
\colorlet{IndoorLight}{IndoorDark!70!white}
\colorlet{SubTLight}{SubTDark!70!white}
\colorlet{AerialLight}{AerialDark!70!white}
\colorlet{UnderWaterLight}{UnderWaterDark!70!white}

\newcommand{\scenario}[1]{{\fontsize{9}{8.7}\selectfont\sf#1}\xspace}
\newcommand{\scenariob}[1]{{\fontsize{14}{14}\selectfont\sf#1}\xspace}
\newcommand{\scenariot}[1]{{\fontsize{8}{8}\selectfont\sf#1}\xspace}
\newcommand{\scenariotn}[1]{{\fontsize{8}{8}\selectfont\sf#1}}

\newcommand{\SVNICP}{\scenariot{SVN-ICP}}
\newcommand{\SVGDICP}{\scenariot{SVGD-ICP}}

\newcommand{\SVNICPn}{\scenariotn{SVN-ICP}}
\newcommand{\sgdICP}{\scenariot{SGD-ICP}}
\newcommand{\steinICP}{\scenariot{Stein ICP}}

\DeclareMathAlphabet\mathbfcal{OMS}{cmsy}{b}{n}

\expandafter\def\expandafter\normalsize\expandafter{%
    \normalsize%
    \setlength\abovedisplayskip{3pt}%
    \setlength\belowdisplayskip{3pt}%
    \setlength\abovedisplayshortskip{-8pt}%
    \setlength\belowdisplayshortskip{2pt}%
}

\title{\LARGE \bf
\scenariob{SVN-ICP}: Uncertainty Estimation of ICP-based LiDAR Odometry \\ using Stein Variational Newton
}

\author{Shiping Ma$^{1\dagger}$,~\IEEEmembership{Student Member,~IEEE}, Haoming Zhang$^{2\dagger}$,~\IEEEmembership{Member,~IEEE}, and Marc Toussaint$^{1}$,~\IEEEmembership{Member,~IEEE}

\thanks{$\dagger$: Equally contributed authors.}
\thanks{$^{1}$Shiping Ma and Marc Toussaint are with the Learning \& Intelligent Systems Research Lab, TU Berlin, Berlin, Germany.} 
\thanks{$^{2}$Haoming Zhang is with the Learning Systems and Robotics Lab, Technical University of Munich, Munich, Germany, and with the Munich Institute of Robotics and Machine Intelligence (MIRMI).} 
\thanks{ Corresponding: \href{mailto:shiping.ma@tu-berlin.de}{shiping.ma@tu-berlin.de} and \href{mailto:haoming.zhang@tum.de}{haoming.zhang@tum.de}.}
}

\begin{document}
\IEEEaftertitletext{\vspace{-.6\baselineskip}}
\maketitle
\markboth{IEEE Robotics and Automation Letters. Preprint Version. Accepted September, 2025}
{Ma \MakeLowercase{\textit{et al.}}: SVN-ICP: Uncertainty Estimation of ICP-based LiDAR Odometry using Stein Variational Newton}

\input{sec/0_abstract}
\input{sec/1_introduction}
\input{sec/2_related_work}
\input{sec/3_preminlaries}
\input{sec/4_SVN-ICP}
\input{sec/5_implementation}
\input{sec/6_results_discussion}

\input{sec/7_conclusion_acknowledgement}

\addtolength{\textheight}{-12cm}

\bibliographystyle{IEEETran.bst}
\bibliography{reference}
\end{document}

%% file: sec/0_abstract.tex
\begin{abstract}
This letter introduces \SVNICP, a novel Iterative Closest Point (ICP) algorithm with uncertainty estimation that leverages Stein Variational Newton (SVN) on manifold. Designed specifically for fusing LiDAR odometry in multisensor systems, the proposed method ensures accurate pose estimation and consistent noise parameter inference, even in LiDAR-degraded environments. By approximating the posterior distribution using particles within the Stein Variational Inference framework, \SVNICP eliminates the need for explicit noise modeling or manual parameter tuning. To evaluate its effectiveness, we integrate \SVNICP into a simple error-state Kalman filter alongside an IMU and test it across multiple datasets spanning diverse environments and robot types. Extensive experimental results demonstrate that our approach outperforms best-in-class methods on challenging scenarios while providing reliable uncertainty estimates. We release our code at \url{https://github.com/LIS-TU-Berlin/SVN-ICP.git}. A high-resolution video demonstration is available at \url{https://youtu.be/c4QsMd1weik}.
\end{abstract}


%% file: sec/1_introduction.tex
\section{Introduction}\label{sec: introduction}

\IEEEPARstart{L}{ocalization} using light detection and ranging (LiDAR) sensors has become a cornerstone of robotic navigation over the past decades. Owing to their high ranging accuracy and robustness to varying lighting conditions, LiDAR sensors produce point clouds that can be exploited through scan matching techniques to estimate robot's pose. 

In general, scan matching in LiDAR odometry (LO) can be performed at the raw point level (i.e.\ direct LO) or at the feature level (i.e.\ feature-based LO) \cite{LO_survey_Lee}. Unlike the latter, direct LO is generally considered more generalizable and accurate at the cost of higher computational complexity, as it conducts scan matching on point level. 

Taken the point-to-point ICP algorithm \cite{icp} as an example, scan matching aims to find the best transformation $\myFrameVecHat{T}{}{} = \{\myFrameVecHat{R}{}{},~\myFrameVecHat{p}{}{}\}\in SO(3)\times\mathbb{R}^3$ between two associated point clouds by solving
\begin{align}
    \myFrameVecHat{T}{}{} \leftarrow \argminD_{\myFrameVec{T}{}{}}\underbrace{\sum_{n=1}^N\normsq{\myFrameVec{q}{n}{}-(\myFrameVec{R}{}{}\myFrameVec{p}{n}{}+\myFrameVec{p}{}{})}{}}_{\text{Loss function}~\mathcal{L}(\myFrameVec{T}{}{})}\label{eq: icp_objective},
\end{align}
where each point $\myFrameVec{p}{n}{}$ in the source point cloud with $N$ points can be transformed by $\myFrameVecHat{T}{}{}$ to match its corresponding point $\myFrameVec{q}{n}{}$ in the target point cloud. 

However, since the optimization in \eqref{eq: icp_objective} yields only a point estimate, it lacks the uncertainty characterization necessary for multisensor fusion.
Many prior fusion methods rely on fixed heuristics \cite{liosam} and hand-crafted noise models based on sensor noise, geometry, or scan convergence \cite{LIO-EKF}.
In contrast, prior studies \cite{censi, icp-cov-martin} highlight two key sources of uncertainty: (a) degenerative observations from noise or poor geometry, and (b) algorithmic issues like local minima or incorrect prior information. Beyond the fact that many of these sources cannot be sufficiently captured in advance, degraded environments, such as dust-filled caves, may exhibit a combination of multiple uncertainty sources, making uncertainty modeling a challenging problem. 

\begin{figure}[!t]
    \centering
    \includegraphics[width=0.48\textwidth]{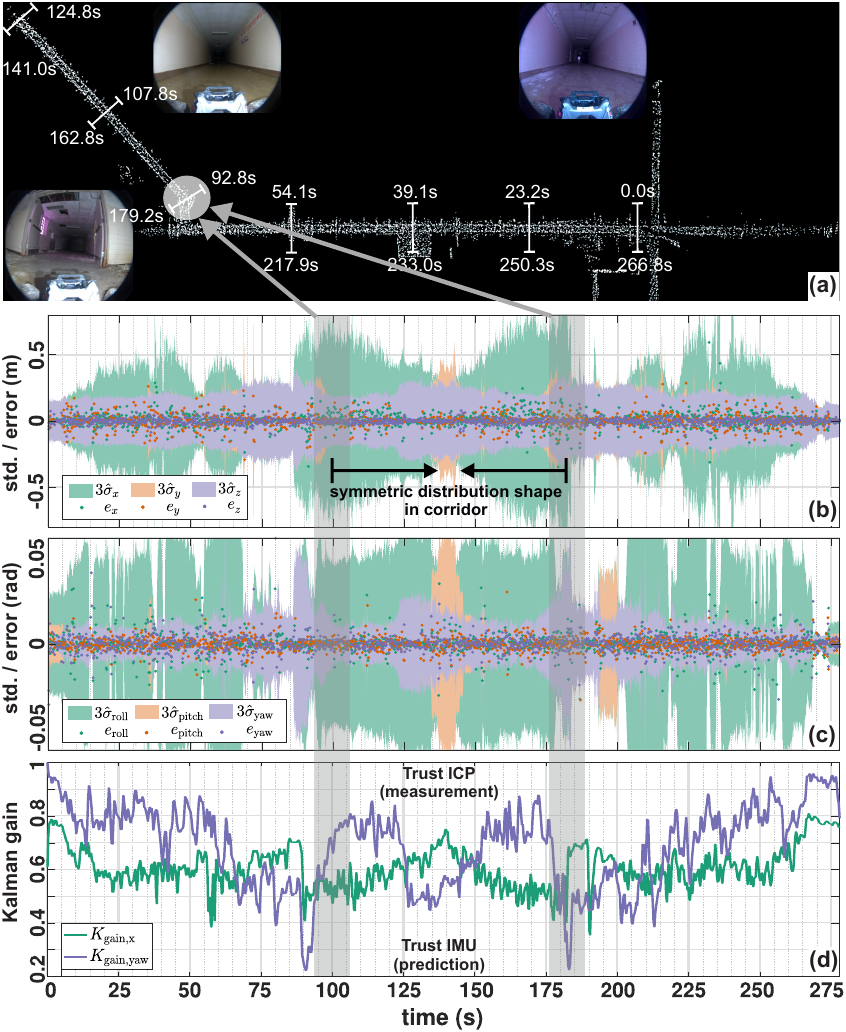}
    \caption{Uncertainty-aware LiDAR-Inertial Odometry using \SVNICP (ours) evaluated on the \texttt{Long Corridor} sequence from the \texttt{SubT-MRS} dataset \cite{Data_subtmrs}. (a) Generated LiDAR map with timestamp annotations. (b-c) Estimated $6$-D uncertainties (variance) visualized as $3\myFrameScalarHat{\sigma}{}{}$ bounds along with error samples from the test run. (d) Smoothed Kalman gain computed given the estimated ICP variance.}
    \label{fig: long_corridor}
\end{figure}

To address this problem, Maken et al. \cite{stein-icp} proposed \steinICP, which advances the probabilistic ICP algorithm by leveraging Stein variational gradient descent (SVGD) \cite{SVGD} to enable principled uncertainty estimation. SVGD is well-known as a parallelizable and practical implementation of Stein variational inference that approximates the posterior distribution using a set of particles. In turn, a direct uncertainty measure can be naturally derived from the non-parametric variational distribution without requiring any explicit uncertainty modeling or detection techniques.

However, \steinICP still faces several limitations. As a first-order method, it inherits the drawbacks of gradient-based updates and performs poorly on ill-conditioned problems \cite{MarteauFerey2019GlobalNewton}, particularly in environments with degenerate structures. This often results in slow, zig-zag convergence and high sensitivity to step size.
Moreover, its Euclidean pose representation requires separate gradient computations and kernel designs for translation and rotation, which can deviate from the underlying $SE(3)$ manifold and hinder convergence.

Motivated by these observations, we propose \SVNICP, a method tailored for LiDAR-based robot localization that extends previous approaches with two key innovations. First, it employs manifold-based state representation for computing scan-matching gradients using the right-hand perturbation model, eliminating the need for separately designing the kernel and computing the gradient for the translational and rotational parts. Second, it replaces SVGD with Stein Variational Newton (SVN) \cite{SVN}, incorporating second-order information (i.e.\ Hessian) to improve convergence stability and speed. Additionally, voxel-based point sampling and an early-stopping criterion further enhance efficiency and robustness.

Our experiments show that these innovations enhance both odometry and sensor fusion performance. Unlike \cite{stein-icp}, we thoroughly evaluate standalone LO and uncertainty estimates in a loosely coupled LiDAR–inertial system using lengthy real-world datasets, with both quantitative and qualitative analysis. Fig.\,\ref{fig: long_corridor} exemplifies the effectiveness of \SVNICP's uncertainty estimation, validated through a Kalman filter, where the Kalman gain reliably reflects LiDAR degradation (e.g.,\ in corridor-like environments). 

Our contributions are threefold: 1) \textbf{Novel Algorithm:} We propose a novel ICP-based LiDAR odometry method, \SVNICP, featuring built-in uncertainty quantification using Stein Variational Newton (SVN) on manifold, 2) \textbf{Comprehensive Evaluation:} We evaluate the estimated uncertainty by integrating \SVNICP into a Kalman filter and benchmarking it against best-in-class methods on two challenging datasets, 3) \textbf{Implementation:} we develop an early-stopping mechanism for SVN and provide an efficient implementation in \texttt{C++} with \texttt{GPU} support. 

%% file: sec/2_related_work.tex
\section{Related Work}

\subsection{LiDAR-based Odometry}
Decades of intensive research have shaped the current state of LiDAR-based odometry, which is primarily categorized into feature-based and direct methods \cite{LO_survey_Lee}. 

Within this scope, feature-based LiDAR odometry relies on matching sparse, distinctive geometric features, such as edges and planar \cite{f-loam}, or surface structures \cite{suma}, and achieves high performance with efficient computation in structured environments. Another prominent class of methods performs scan matching directly at the raw point cloud level, typically building on the Iterative Closest Point (ICP) \cite{icp} or Normal Distribution Transform (NDT) \cite{ndt} algorithms. Since ICP-based methods generally achieve higher accuracy \cite{icp_vs_ndt}, many recent studies have advanced this technique by leveraging temporal continuity in scan matching \cite{ct-icp},
adaptively adjusting thresholding parameters \cite{kiss-icp}, or maintaining map representations effectively by considering structural information and estimation uncertainty \cite{mad-icp}.

Often, scan matching is integrated as a measurement update within a state estimator (e.g.,\ a Kalman filter), with state propagation performed using an Inertial Measurement Unit (IMU), resulting in a LiDAR-inertial odometry system. Most notable approaches perform tightly coupled fusion using Kalman filters, where residuals of individual feature points are computed to correct the propagated state \cite{FastLio2, LIO-EKF}, achieving both high accuracy and robustness.

However, since these methods primarily target nominal operating conditions, their performance generally degrades in challenging environments. To address this issue, many recent studies leverage additional incidental measurements from LiDAR, such as intensity \cite{intensity_2, coin-lio} and Doppler velocity \cite{yuchen_lo_Doppler}, to enhance robustness in degraded scenarios.

\textit{\textbf{Discussion:}} Despite their success in robot localization, existing methods do not explicitly address uncertainty awareness in degeneracy scenarios. Most rely on hand-crafted noise models or additional inputs, lacking reliable uncertainty estimates that are crucial for robust state estimation. 

\subsection{Uncertainty-Aware LiDAR Odometry}
Recent studies have increasingly focused on incorporating uncertainty awareness into LiDAR-based localization, which can be broadly divided into degeneracy detection and uncertainty estimation.

Degeneracy (aka\ localizability) detection adopts a discriminative perspective by analyzing normal vectors \cite{la_lio}, scan-matching residuals \cite{genz-icp}, or the Hessian matrix \cite{x-icp, johan_ral}, to down-weight degraded features or directions and reshape the covariance matrix for optimization. Recent approaches also extend this concept by incorporating multiple LiDARs \cite{multilidar_Localizability} or additional sensing modalities \cite{RELEAD, Deg_LidarRadar, SuperLoc} to better handle LiDAR degradation. Furthermore, learning-based methods for localizability detection in LiDAR point clouds have been investigated \cite{learning_localizability_Julian}. 

In contrast to degeneracy detection, the second line of work focuses on directly quantifying the uncertainty of ICP-based LiDAR odometry. While classical ICP methods using single value decomposition \cite{icp_cov_svd} or quaternions \cite{icp} can compute covariance matrices, many earlier studies have noted their tendency to be overoptimistic and proposed closed-form covariance estimations by explicitly modeling key sources of uncertainty \cite{censi, icp-cov-martin}, or by adopting learning-based approaches \cite{icp-cov-learning,icp_deep_bayesian} to better capture the uncertainty in complex environments. 

Another principled scan-matching approach with uncertainty estimation is based on the Bayesian formulation, exemplified by sampling-based Bayesian ICP \cite{Bayesian-ICP-full} which uses a computation-intensive Markov Chain Monte Carlo (MCMC) algorithm to approximate the distribution of the estimated transformation. To improve efficiency, the same authors later proposed \steinICP \cite{stein-icp}, which replaces MCMC with Stein variational inference, significantly reducing computational cost while maintaining comparable accuracy.

\textit{\textbf{Discussion:}} While methods based on degeneracy detection either require carefully tuned threshold parameters or rely on additional sensor inputs within complex frameworks, they still face challenges in threshold tuning and system adaptation across diverse environments. Other approaches that explicitly model or implicitly predict sources of uncertainty may struggle to generalize across scenarios, often resulting in overconfident uncertainty estimates. 

These limitations motivate our investigation of uncertainty estimation for LiDAR odometry using Stein variational inference, a one-shot method to approximate the ICP posterior that avoids extensive tuning and over-engineering.

%% file: sec/3_preminlaries.tex
\section{Preliminaries}\label{sec: prelimiaries}
\subsection{Stein Variational Gradient Descent}\label{sec: pre_svgd}
Stein's method \cite{SVGD} establishes a probabilistic distance measure between the posterior (target) distribution $p(\myFrameVec{\xi}{}{})$ and a latent distribution $q(\myFrameVec{\xi}{}{})$ by computing the kernelized Stein discrepancy

{\small
\begin{align}
\label{eq: stein_discrepancy}
     \mathop{\mathbb{S}}(q,p) = \max_{\myFrameVec{\phi}{}{}\in\mathcal{H}^d}\left\{[\mathbb{E}_{\myFrameVec{\xi}{}{}\sim q}\mathrm{Tr}(\mathcal{A}_p \myFrameVec{\phi}{}{}(\myFrameVec{\xi}{}{}))]^2,~s.t.~\norm{\myFrameVec{\phi}{}{}}_{\mathcal{H}^d}\leq1\right\}
\end{align}
}using a vector-valued smooth function $\myFrameVec{\phi}{}{}(\myFrameVec{\xi}{}{})=[\myFrameScalar{\phi}{1}{}(\myFrameVec{\xi}{}{}),\cdots,\myFrameScalar{\phi}{d}{}(\myFrameVec{\xi}{}{})]$ from a reproducing kernel Hilbert space (RKHS) $\mathcal{H}^d$. The function $\mathrm{Tr}(\cdot)$ computes the matrix trace. And the function  $\myFrameVec{\phi}{}{}(\myFrameVec{\xi}{}{})$ must satisfy the Stein identity, i.e., $\mathbb{E}_{\myFrameVec{\xi}{}{}\sim p}[\mathcal{A}_p \myFrameVec{\phi}{}{}(\myFrameVec{\xi}{}{})]=0$, where the Stein operator is defined as
\begin{align}
\label{eq: stein_operator}
    \mathcal{A}_p\myFrameVec{\phi}{}{}(\myFrameVec{\xi}{}{}) = \nabla_{\myFrameVec{\xi}{}{}}\log p(\myFrameVec{\xi}{}{})^T + \nabla_{\myFrameVec{\xi}{}{}}\myFrameVec{\phi}{}{}(\myFrameVec{\xi}{}{}).
\end{align}
In particular, the quantity $\mathbb{E}_{\myFrameVec{\xi}{}{}\sim q}[\mathrm{Tr}(\mathcal{A}_p \myFrameVec{\phi}{}{}(\myFrameVec{\xi}{}{}))]^2$ serves as a discrepancy metric that equals zero if and only if $q = p$.

Taken the Stein discrepancy as a variational objective, SVGD \cite{SVGD} employs a set of $K$ particles $\{\myFrameVec{\xi}{k}{}\}_{k}^K$ to represent $q$, updated with a small perturbation $\epsilon\myFrameVec{\phi}{}{}(\myFrameVec{\xi}{k}{})$ as
\begin{align}
\label{eq: stein_state_update}
    \myFrameVec{\xi}{k}{} \leftarrow \myFrameVec{\xi}{k}{} + \epsilon\myFrameVec{\phi}{}{}(\myFrameVec{\xi}{k}{}),~~~k = 1,\dots,K,
\end{align}
which gradually drives the particles to match the target distribution $p$. The optimal vector function $\myFrameVec{\phi}{}{*}$, in the limit as the step size $\epsilon\rightarrow0$, corresponds to the steepest descent direction of the Kullback-Leibler (KL) divergence, which gives us
\begin{align}
\begin{split}
\label{eq: optimal_phi}
    \myFrameVec{\phi}{}{*} &= \arg\max_{\myFrameVec{\phi}{}{}} \left\{-\frac{\mathrm{d}}{\mathrm{d}\epsilon}KL(q_{\myFrameVec{\phi}{}{}} \,\Vert\, p)\big|_{\epsilon=0}\right\}. 
\end{split}
\end{align}

As all values in the function $\myFrameVec{\phi}{}{}$ in \eqref{eq: stein_discrepancy} lie within the unit ball $\mathcal{B}=\{\myFrameVec{\phi}{}{}\in\mathcal{H}^d:~\norm{\myFrameVec{\phi}{}{}}_{\mathcal{H}^d}\leq 1\}$ of $\mathcal{H}^d$ in the vector-valued RKHS, defined by the positive definite kernel $k(\myFrameVec{\xi}{}{}, \cdot)$, we can exploit the reproducing property $\mathbb{E}_{q}[\mathcal{A}_p \myFrameVec{\phi}{}{}(\myFrameVec{\xi}{}{})] = \langle\bm{\phi},~\mathbb{E}_q[\mathcal{A}_pk(\myFrameVec{\xi}{}{}, \cdot))] \rangle$ to compute \eqref{eq: optimal_phi}.
By substituting \eqref{eq: stein_operator} in \eqref{eq: optimal_phi}, we have
\begin{align}
    \myFrameVec{\phi}{}{*} &= \mathbb{E}_{\myFrameVec{\xi}{}{}\sim q}\mathrm{Tr}(\mathcal{A}_p k(\myFrameVec{\xi}{}{}, \cdot))\\
    &= \mathbb{E}_{\myFrameVec{\xi}{}{}\sim q}[\underbrace{\nabla_{\myFrameVec{\xi}{i}{}}\log p(\myFrameVec{\xi}{i}{})k(\myFrameVec{\xi}{}{},\cdot)}_{\text{Steepest descent direction}} + \underbrace{\nabla_{\myFrameVec{\xi}{i}{}}k(\myFrameVec{\xi}{}{},\cdot)}_{\text{Repulsion force}}],\label{eq: SVGD_update}
\end{align}
where the first term directs particles toward regions of high probability under $p(\myFrameVec{\xi}{i}{})$, and the second term introduces a repulsive force that discourages particles from collapsing around a single mode, thereby promoting spread across the support of $p(\myFrameVec{\xi}{i}{})$.

\subsection{\scenario{Stein ICP}}\label{sec: pre_stein_icp}
\steinICP \cite{stein-icp} follows the concept of the \sgdICP \cite{SGD-ICP}, which solves the ICP problem in \eqref{eq: icp_objective} using stochastic gradient descent \cite{SGD}. In contrast to standard ICP solvers, \sgdICP partitions the dense point cloud into mini-batches $\mathbfcal{M}=\{\mathcal{M}_1\dots\mathcal{M}_M\}$ and optimizes the transformation $\myFrameVecHat{T}{ij}{}$ by iteratively updating the state parameters, similarly to \eqref{eq: stein_state_update}. However, unlike \eqref{eq: stein_state_update}, the gradient $\myFrameScalarBar{g}{}{}$ of the state parameter $\myFrameVec{\xi}{}{}$ must be separately computed for the translational and the rotational parts (see Equation (4)-(5) in \cite{stein-icp}).

Built on \sgdICP, \steinICP leverages SVGD to compute gradients of the state parameters in \eqref{eq: stein_state_update} by adapting \eqref{eq: SVGD_update} as 

{
\smaller
\begin{align}
\label{eq: SteinICP_update}
\begin{split}
    \myFrameVec{\phi}{}{*} = \sum_{k=1}^K\Bigl[ -\bigl(\underbrace{N\myFrameScalarBar{g}{}{}(\myFrameVec{\xi}{k}{t},\mathcal{M}^t)+ \nabla\log p_{0}(\myFrameVec{\xi}{k}{})}_{\nabla_{\myFrameVec{\xi}{k}{}}\log p(\myFrameVec{\xi}{k}{})}\bigr) k(\myFrameVec{\xi}{k}{},\myFrameVec{\xi}{}{}) + \nabla_{\myFrameVec{\xi}{k}{}}k(\myFrameVec{\xi}{k}{},\myFrameVec{\xi}{}{}) \Bigr],
\end{split}
\end{align}}where the gradient of the posterior log-density $\nabla_{\myFrameVec{\xi}{k}{}}\log p(\myFrameVec{\xi}{k}{})$ consists of two components: the mini-batch gradients computed, and the gradient of the log prior distribution (see Equation (14)-(15) in \cite{stein-icp}). An RBF kernel with a modified distance metric is used.  

%% file: sec/4_SVN-ICP.tex
\section{\scenario{SVN-ICP}} \label{sec: stein_micp}

To overcome the limitations of \steinICP introduced in Sec.\,\ref{sec: introduction}, we propose \SVNICP, a frame-to-map LiDAR odometry method with uncertainty estimation that improves upon \steinICP in both accuracy and convergence speed. 

\subsection{ICP on Manifold}

In this work, we consider solving the point-to-point scan matching problem in \eqref{eq: icp_objective} by formulating the state transformation on a manifold using the following loss function:
\begin{align}
    \mathcal{L}(\myFrameVec{\xi}{}{}) = \sum_{n=1}^{N}\left\|\right.\underbrace{(\exp{(\myFrameVec{\vartheta}{}{})}\myFrameVec{p}{n}{}+\myFrameVec{p}{}{}) -  \myFrameVec{q}{n}{}}_{\myFrameVec{e}{n}{}}\left.\right\|^2,\label{eq: icp_objective_se3}
\end{align}
where the vector $\myFrameVec{\xi}{}{} = [\myFrameVec{p}{}{}~\myFrameVec{\vartheta}{}{}]^T =[x~y~z~\myFrameScalar{\theta}{x}{}~\myFrameScalar{\theta}{y}{}~\myFrameScalar{\theta}{z}{}]^T \in\mathbb{R}^3\times\mathfrak{so}(3)$ denotes the minimal pose increment between two point clouds. 
To enable iterative optimization, we introduce the right-hand perturbation state $\Delta\myFrameVec{\xi}{}{}$ around a nominal state $\myFrameVecBar{\xi}{\rm op}{}$ in \eqref{eq: icp_objective_se3}. This yields the perturbed loss function:
\begin{align}
    \mathcal{L}(\myFrameVecBar{\xi}{\rm op}{} \boxplus \Delta\myFrameVec{\xi}{}{}) = \sum_{n=1}^{N}\myFrameVec{e}{n}{}(\myFrameVecBar{\xi}{\rm op}{} \boxplus \Delta\myFrameVec{\xi}{}{})^{T}\myFrameVec{e}{n}{}(\myFrameVecBar{\xi}{\rm op}{} \boxplus \Delta\myFrameVec{\xi}{}{}), \label{eq: icp_objective_se3_perturbation}
\end{align}
where $\boxplus$ denotes the pose update on manifold \cite{blancoclaraco2022tutorial}. 

As \SVNICP follows Newton's method to optimize \eqref{eq: icp_objective_se3_perturbation}, we approximate the loss function using the second-order Taylor expansion as
\begin{align}
    \mathcal{L}(\myFrameVecBar{\xi}{\rm op}{}\boxplus\Delta\myFrameVec{\xi}{}{})\approx\mathcal{L}(\myFrameVecBar{\xi}{\rm op}{}) + 2\myFrameVec{b}{}{T}\Delta\myFrameVec{\xi}{}{} + \Delta\myFrameVec{\xi}{}{T}\myFrameVec{H}{}{}\Delta\myFrameVec{\xi}{}{},\label{eq: ICP_Hessian}
\end{align}
where the gradient $\myFrameVec{b}{}{}$ and the approximated Hessian $\myFrameVec{H}{}{}$ are given by
\begin{align} 
\myFrameVec{b}{}{} = \sum_{n=1}^{N}\myFrameVec{b}{n}{}, ~~~ \myFrameVec{H}{}{} = \sum_{n=1}^{N}\myFrameVec{H}{n}{}\label{eq: ICP_b_H}.
\end{align}
The Jacobian $\myFrameVec{J}{n}{}$, gradient $\myFrameVec{b}{n}{}$, and Hessian $\myFrameVec{H}{n}{}$ of the point $\myFrameVec{p}{n}{}$ in the ICP process are computed as
\begin{align}
\begin{split}
    \myFrameVec{J}{n}{} &=\frac{\partial\myFrameVec{e}{n}{}}{\partial\myFrameVec{\xi}{}{}} = \left[ -\left[\myFrameVec{R}{}{}\myFrameVec{p}{n}{}\right]_{\times} ~ \myFrameVec{R}{}{}\right]\in\mathbb{R}^{3\times6}, 
    \\ \myFrameVec{b}{n}{} &= -\myFrameVec{J}{n}{T}\myFrameVec{e}{n}{},~~~\myFrameVec{H}{n}{} = \myFrameVec{J}{n}{T}\myFrameVec{J}{n}{},
\end{split}\label{eq: right_jacobian_hessian}
\end{align}
where the operator $[\cdot]_{\times}$ converts the $3$-D vector into its corresponding skew-symmetric matrix.

In this scheme, the optimal perturbation state $\Delta\myFrameVec{\xi}{}{}$ is obtained by solving the linear system $\myFrameVec{H}{}{}\Delta\myFrameVec{\xi}{}{} = \myFrameVec{b}{}{}$, which forms the estimated transformation given as
\begin{equation}
    \label{eq: right_perturbation}
    \myFrameVecHat{T}{}{} = \myFrameVecBar{T}{\rm op}{}\mathrm{exp}(\Delta\myFrameVec{\xi}{}{}).
\end{equation}

Unlike prior works that use a left-hand perturbation model \cite{kiss-icp, LIO-EKF, genz-icp}, we adopt a right-hand formulation that better decouples translation and rotation. This mitigates the coupling effects inherent in left-multiplicative updates and improves numerical stability in frame-to-map ICP, especially under large initial translations or when handling distant points, conditions frequently seen in odometry applications.

\subsection{Solving ICP with Stein Variational Newton}
We employ variational inference within the probabilistic ICP framework using the Stein Variational Newton (SVN) method \cite{SVN}. In contrast to SVGD, which follows the gradient flow of the KL divergence (cf. \eqref{eq: optimal_phi}), SVN incorporates second-order information of the objective function \eqref{eq: icp_objective_se3_perturbation} to compute Newton-like updates in function space.

Concretely, SVN seeks a Newton-like update direction under the approximation
\begin{equation}
    -\frac{\mathrm{d}}{\mathrm{d}\epsilon}KL(q_{\myFrameVec{\phi}{}{}}\Vert p)\big|_{\epsilon=0} \approx \frac{\mathrm{d}^2}{\mathrm{d}\epsilon^2}KL(q_{\myFrameVec{[\phi, \theta]}{}{}}\Vert p)\big|_{\epsilon=0},
\end{equation}
where $\myFrameVec{\phi}{}{}$ is an arbitrary first-order functional perturbation (as in \eqref{eq: optimal_phi}) and 
$\myFrameVec{\theta}{}{}$ denotes an independent second-order perturbation. Accordingly, we define the Stein variational update of $k$-th particle with step size $\epsilon=1$ as in \cite{SVN} as 
\begin{align}
\myFrameVec{\xi}{k}{} \leftarrow \myFrameVec{\xi}{k}{} + \epsilon\myFrameVec{\theta}{}{*}(\myFrameVec{\xi}{k}{}),  ~~~\myFrameVec{\theta}{}{*} =  \tilde{\myFrameVec{H}{}{}}^{-1}\myFrameVec{\phi}{}{*}
\end{align}
with the optimal Newton direction $\myFrameVec{\theta}{}{*}$. Following \steinICP (Sec.\,\ref{sec: pre_stein_icp}), we compute the SVGD update $\myFrameVec{\phi}{}{*}$ in \eqref{eq: SVGD_update} as
\begin{align}
    \myFrameVec{\phi}{}{*}(\myFrameVec{\xi}{k}{})=\frac{1}{K}\sum_{l=1}^{K}\left[  k(\myFrameVec{\xi}{l}{},\myFrameVec{\xi}{k}{})\myFrameVec{b}{}{}(\myFrameVec{\xi}{k}{}) + \nabla_{\myFrameVec{\xi}{l}{}}k(\myFrameVec{\xi}{l}{},\myFrameVec{\xi}{k}{}) \right] \label{eq: svn_svgd_update}.
\end{align}

The matrix-preconditioned Hessian $\myFrameVecTilde{H}{}{}$ of SVN associated with the particle $\myFrameVec{\xi}{k}{}$ is given by 

{\small
\begin{align}
    \myFrameVecTilde{H}{}{}(\myFrameVec{\xi}{k}{}) &= \mathbb{E}_{\myFrameVec{\xi}{}{}\sim q} \left[ \myFrameVec{H}{k}{}(\myFrameVec{\xi}{}{})k(\myFrameVec{\xi}{}{}, \myFrameVec{\xi}{k}{})^2 + (\nabla_{\myFrameVec{\xi}{}{}}k(\myFrameVec{\xi}{}{}, \myFrameVec{\xi}{k}{}))^{\otimes2}\right] \nonumber \\
    &= \frac{1}{K}\sum_{l=1}^{K}\left[ \myFrameVec{H}{k}{}(\myFrameVec{\xi}{}{})k(\myFrameVec{\xi}{l}{}, \myFrameVec{\xi}{k}{})^2 + (\nabla_{\myFrameVec{\xi}{}{}}k(\myFrameVec{\xi}{l}{}, \myFrameVec{\xi}{k}{}))^{\otimes2}\right]\label{eq: SVN_Hessian},
\end{align}}where the repulsive force is computed using the operation $\myFrameVec{w}{}{\otimes2}=\myFrameVec{w}{}{}\myFrameVec{w}{}{\top}$. 

In \eqref{eq: svn_svgd_update} and \eqref{eq: SVN_Hessian}, we use the RBF kernel 
   $ k(\myFrameVec{\xi}{l}{}, \myFrameVec{\xi}{k}{}) = \exp{\left(\frac{1}{h}\normsq{\myFrameVec{\xi}{l}{}-\myFrameVec{\xi}{k}{}}{2}\right)}$
with bandwidth $h$ in Stein's methods following \cite{stein-icp}.

\renewcommand{\algorithmiccomment}[1]{/* \textit{#1} */} 
\SetKwInOut{KwIn}{Input}
\SetKwInOut{KwOut}{Output}
\SetKwBlock{Pfor}{parallelized for}{end}
\begin{algorithm}[!t] 
\KwIn{Prior pose $\myFrameVecCheck{T}{}{}$ and covariance $\myFrameVecCheck{\Sigma}{}{}$, \\
      Source cloud $\mathbfcal{P} = \{\myFrameVec{p}{n}{}\},~ n=1\dots N_p$, \\ 
      Target cloud $\mathbfcal{Q}  = \{\myFrameVec{q}{n}{}\},~ n=1\dots N_q$.}
     
\KwOut{Posterior pose $\myFrameVecHat{T}{}{}$ and covariance $\myFrameVecHat{\Sigma}{}{}$}
\caption{\SVNICP} \label{alg: stein-micp}
\tcc{\scriptsize initialize $K$ perturbation particles}
$\myFrameVec{\Xi}{}{}=\{\myFrameVec{\xi}{k}{}\}_{1}^{K},~~\text{where}~\myFrameVec{\xi}{k}{} \sim \mathcal{N}(\myFrameVec{0}{}{}, \myFrameVec{\sigma}{}{})$ ; \\

\tcc{\scriptsize find corresponding sub target point cloud $\mathbfcal{Q}_n$} 
\Pfor(each point $\myFrameVec{p}{n}{} \in \mathbfcal{P}$){ 
$ \mathbfcal{Q}_{n} \leftarrow \mathrm{KNN}(\myFrameVec{p}{n}{}, \mathbfcal{Q})$ ;
}

\For{iteration $i=1\dots I$ }{
\Pfor(each particle $\myFrameVec{\xi}{k}{} \in \myFrameVec{\Xi}{}{}$)  {
    {\small
        \tcc{\scriptsize update pose in new iteration}
        $\myFrameVec{T}{k, i}{} = \{\myFrameVec{R}{k, i}{},~\myFrameVec{p}{k, i}{}\} = \myFrameVecCheck{T}{}{}\myFrameVec{T}{k, i-1}{}$ ;\\
        \Pfor(each point $\myFrameVec{p}{n}{} \in \mathbfcal{P}$){
        $\myFrameVec{p}{n}{\prime} = \myFrameVec{R}{k,i}{}\myFrameVec{p}{n}{} + \myFrameVec{p}{k,i}{}$ \tcp*{\scriptsize transform point}
        $\myFrameVec{q}{n}{} \leftarrow \mathrm{KNN}(\myFrameVec{p}{n}{\prime}, \mathbfcal{Q}_{n})$ \tcp*{\scriptsize find correspond}
        Compute point residual $\myFrameVec{e}{n,k}{} = \myFrameVec{p}{n}{\prime} - \myFrameVec{q}{n}{}$ ;\\
        Compute $\myFrameVec{J}{n,k}{}$, $\myFrameVec{b}{n,k}{}$, $\myFrameVec{H}{n, k}{}$ \tcp*{\scriptsize Equation \eqref{eq: right_jacobian_hessian}}
        }
        Compute $\myFrameVec{b}{k,i}{}$ and $\myFrameVec{H}{k,i}{}$ \tcp*{\scriptsize Equation \eqref{eq: ICP_b_H}}
        Compute SVN Hessian $\myFrameVecTilde{H}{k,i}{}$ \tcp*{\scriptsize Equation \eqref{eq: SVN_Hessian}}
        Compute state update $\Delta\myFrameVec{\xi}{k,i}{}={\myFrameVecTilde{H}{k,i}{-1}}\myFrameVec{\phi}{k,i}{*}$ ; \\ 
        Update particle pose $\myFrameVec{T}{k,i}{} = \myFrameVecHat{T}{k}{}\exp{(\Delta\myFrameVec{\xi}{k}{})}$ ;\\ 
    }
 }
 \If{$\frac{1}{K}\sum_{k=1}^{K}\left\| \Delta\myFrameVec{\xi}{k}{} \right\|^2 < \myFrameVec{\epsilon}{}{}$}{
            break \tcp*{\scriptsize early-stop condition reached}
          }
}
\tcc{\scriptsize compute mean perturbation state and covariance}
{\smaller
$\myFrameVecBar{\xi}{\rm icp}{}= \frac{1}{K}\sum_{k=1}^{K}\myFrameVec{\xi}{k}{}, ~~~\myFrameVecBar{\Sigma}{\rm icp}{}=\frac{1}{K}\sum_{k=1}^{K}\left( \myFrameVec{\xi}{k}{}-\myFrameVecBar{\xi}{}{} \right) \left( \myFrameVec{\xi}{k}{}-\myFrameVecBar{\xi}{}{} \right)^{T}$ \label{algo: line25} \\
}
\tcc{\scriptsize compute final pose and propagate covariance}
{\small
$\myFrameVecHat{T}{}{}=\myFrameVecCheck{T}{}{}\exp{(\myFrameVecBar{\xi}{\rm icp}{})}$,~~~$\myFrameVecHat{\Sigma}{}{}=\myFrameVecCheck{\Sigma}{}{}+\mathrm{Ad}_{\myFrameVecCheck{T}{}{}}\myFrameVecBar{\Sigma}{\rm icp}{}\mathrm{Ad}_{\myFrameVecCheck{T}{}{}}^{T}$
}

\Return{$ \myFrameVecHat{T}{}{},~\myFrameVecHat{\Sigma}{}{}$}
\end{algorithm}

\subsection{The \SVNICP Algorithm}
Algorithm \ref{alg: stein-micp} outlines the pipeline of \SVNICP (odometry only). Starting from a prior pose $\myFrameVecCheck{T}{}{}$ with covariance $\myFrameVecCheck{\Sigma}{}{}$, we sample $K$ particles using a predefined variance $\sigma$ to align the source point cloud $\mathbfcal{P}$ to the target $\mathbfcal{Q}$. To account for incremental LiDAR pose changes in odometry, we downsample $\mathbfcal{Q}$ to a sub-target point cloud $\mathbfcal{Q}_n$ via the $\mathrm{KNN}$ algorithm, retaining only neighbors of points in $\mathbfcal{P}$. Unlike the mini-batch sampling in \steinICP, our method ensures higher variational inference quality while significantly reducing computational cost.

After each SVN iteration, we compute the average norm of state updates across all particles. If it falls below the threshold $\myFrameVec{\epsilon}{}{}$, the optimization stops early (line 19–21). The final pose is given by the mean perturbation across particles (see \eqref{eq: right_perturbation}), while the estimated covariance is transformed to the global frame using the Adjoint of the prior pose, $\mathrm{Ad}_{\myFrameVecCheck{T}{}{}}$ \cite{BarfootTRO2014}.

%% file: sec/5_implementation.tex
\section{Implementation}

We evaluate \SVNICP in two settings: standalone LiDAR odometry (LO) for baseline performance validation, and a loosely coupled LiDAR–Inertial odometry (LIO) system using an error-state Kalman filter. We assume that the state propagation in the standalone LO follows a constant-velocity motion model, as also utilized in \cite{kiss-icp, stein-icp}. In the LIO setup, the LiDAR noise parameter is either fixed or adaptively updated using \SVNICPn’s estimated covariance, demonstrating the benefit of its uncertainty estimation.

The Kalman filter state and its error-state vector are defined at the IMU center as
\begin{align}
    \myFrameVec{x}{b}{w} &\coloneqq \{\myFrameVec{T}{b}{w}, ~ \myFrameVec{v}{b}{w}, ~\myFrameVec{b}{\rm acc}{}, ~\myFrameVec{b}{\rm gyro}{}\} \in SE(3)\times\mathbb{R}^{9},\\
    \Delta\myFrameVec{x}{}{} &\coloneqq [\Delta\myFrameVec{p}{}{T} ~ \Delta\myFrameVec{v}{}{T} ~ \Delta\myFrameVec{\vartheta}{}{T} ~ \Delta\myFrameVec{b}{\rm acc}{T} ~ \Delta\myFrameVec{b}{\rm gyro}{T}]^T \in \mathbb{R}^{15},
\end{align}
where $\myFrameVec{T}{b}{w} = \{\myFrameVec{R}{b}{w}~\myFrameVec{p}{b}{w}\}$ and $\myFrameVec{v}{b}{w}$ denote the pose and velocity of the body frame $(\cdot)_b$ with respect to the world frame $(\cdot)^w$. The vectors $\myFrameVec{b}{\rm acc}{}$ and $\myFrameVec{b}{\rm gyro}{}$ represent the accelerometer and gyroscope biases. The orientation error $\Delta\myFrameVec{\vartheta}{}{}=[\Delta\myFrameScalar{\theta}{x}{}~\Delta\myFrameScalar{\theta}{y}{}~\Delta\myFrameScalar{\theta}{z}{}]^T$ is expressed in axis-angle form. 

We adopt the same IMU mechanization from \cite{LIO-EKF} to propagate the state $\myFrameVecCheck{T}{}{}$ and covariance $\myFrameVecCheck{\Sigma}{}{}$. Since \SVNICP provides the optimal perturbation state (i.e.,\ Kalman innovation) $\myFrameVecBar{\xi}{}{}$ and its covariance $\myFrameVecBar{\Sigma}{\rm icp}{}$ (line 23 of Alg.\,\ref{alg: stein-micp}), the error-state vector is computed with Kalman gain $\myFrameVec{K}{\rm g}{}$ as follows
\begin{align}
    \Delta\myFrameVec{x}{}{} = \myFrameVec{K}{\rm g}{} \myFrameVecBar{\xi}{}{},~~\myFrameVec{K}{\rm g}{} = \myFrameVecCheck{\Sigma}{}{}\myFrameVec{C}{}{T}\cdot(\myFrameVec{C}{}{}\myFrameVecCheck{\Sigma}{}{}\myFrameVec{C}{}{T}+\myFrameVecBar{\Sigma}{\rm icp}{})^{-1},
\end{align}
where the observation matrix $\myFrameVec{C}{}{} \in \mathbb{R}^{6\times15}$ is defined as 
\begin{align}
     \myFrameVec{C}{}{} = \begin{bmatrix}
             \mathbf{I}_{3\times3} &\mathbf{0}_{3\times3} &\mathbf{0}_{3\times3} &\mathbf{0}_{3\times3} &\mathbf{0}_{3\times3} \\    
             \mathbf{0}_{3\times3} &\mathbf{I}_{3\times3} &\mathbf{0}_{3\times3} &\mathbf{0}_{3\times3} &\mathbf{0}_{3\times3}
    \end{bmatrix}.
\end{align}

We implemented our approach in \texttt{C++}, using sensor interfaces from the Robot Operating System \texttt{ROS2} \cite{ROS2}. GPU-based data handling and basic math operations were supported by \texttt{LibTorch} \cite{pytorch}.

%% file: sec/6_results_discussion.tex
\input{tables/make_subtmrs_table}
\input{tables/make_geode_table}
\input{tables/make_ablation_table}

\input{tables/make_KL_NNE_table}

\section{Experiments and Results}
To evaluate both odometry performance and the estimated uncertainty, we use two public datasets: \textit{SubT-MRS} \cite{Data_subtmrs} and \textit{GEODE} \cite{Data_geode}, which encompass representative LiDAR-degraded scenarios. While the \textit{SubT-MRS} dataset primarily features mixed indoor environments, the \textit{GEODE} dataset additionally includes challenging outdoor scenarios such as off-road trails and waterways.

We have re-implemented\footnote{[Online] https://github.com/msp666/SteinICP-Odometry-and-Mapping.} the \steinICP based on its original implementation\footnote{[Online] https://bitbucket.org/fafz/stein-icp.}. Since the original \steinICP performs frame-to-frame matching, we provide an enhanced version, \SVGDICP. This enhanced variant follows the same odometry pipeline as \SVNICP but employs SVGD as the optimization scheme. All experiments are conducted on a desktop equipped with an Intel i7-8086K CPU, \SI{32}{GB} of RAM, and an NVIDIA GTX 1080 Ti GPU. 
\makeSubTMRSTable
\subsection{General Performance Metrics}
General error metrics, including the Absolute Pose Error (APE) and Relative Pose Error (RPE), are computed using the \texttt{evo} package \cite{evo}. 

\subsubsection{Results on \texttt{SubT-MRS} Dataset}\label{sec: res_subt_mrs}
We compare \SVNICP in different settings against classical baselines. Table \ref{tab:ATE_RPE_SUBTMRS} presents the performance metrics on the \texttt{SubT-MRS} dataset. 

While most well-engineered systems with loop-closure constraints, originally developed for competition purposes, achieve the highest accuracy on several sequences, our loosely coupled Kalman filter, leveraging the estimated ICP uncertainty, delivers competitive results without additional observations (e.g., loop-closure constraints).
In contrast to other LiDAR-only odometry methods, which often fail in challenging scenarios, the proposed ICP variant optimized via both SVGD and SVN achieves competitive accuracy and demonstrates strong robustness across all test sequences.
By comparison, the original \steinICP \cite{stein-icp}, designed for frame-to-frame ICP using mini-batch sampling, is not suitable for odometry applications in such demanding environments.

When comparing \SVGDICP and \SVNICP, we observe that both methods exhibit similar performance in well-structured yet geometrically degraded environments. This suggests that SVGD, when solved using the Adam optimizer \cite{adam}, provides sufficiently reliable descent directions for effective convergence. However, in mixed-degeneration environments, \SVNICP consistently outperforms \SVGDICP, highlighting the advantage of incorporating curvature information and stable Hessian-based updates in \SVNICP, features that are better suited to handling degenerate conditions (more details in Sec.\,\ref{sec: ab_svgd_svn}). 

When comparing LiDAR-only and LiDAR–Inertial Odometry methods, their performance differences are highly dependent on motion characteristics in our experiments. LiDAR-only approaches can achieve comparable performance in scenarios involving ground vehicles (e.g.,\ UGVs or RC cars) with smooth motion. However, under aggressive motions, encountered with drones or hand-held devices, integrating IMU measurements in a fusion framework significantly enhances the stability and accuracy of the odometry.

Finally, our key result highlights the importance of consistent uncertainty parameterization in sensor fusion. While the Kalman filter configured with fixed ICP noise parameters\footnote{We adopt commonly used noise parameters for LiDAR odometry: $\myFrameVec{\eta}{\rm ICP,p}{} \sim \mathcal{N}(\myFrameVec{0}{}{},1e^{-4})$ and $\myFrameVec{\eta}{\rm ICP,rot}{} \sim \mathcal{N}(\myFrameVec{0}{}{},1e^{-5})$.} fails to produce reliable odometry estimates in several sequences (marked in red), dynamically updating the ICP noise using the proposed \SVNICP leads to accurate and robust state estimation across all tested scenarios.

\subsubsection{Results on \texttt{GEODE} Dataset}\label{sec: res_geode}
\makeGEODETable
Consistent results as in Sec.\,\ref{sec: res_subt_mrs} are also observed on the \texttt{GEODE} dataset, as shown in Table \ref{tab:ATE_RPE_GEODE}. However, since the test scenarios are less challenging compared to those in the \texttt{SubT-MRS} dataset, the noise-updated Kalman filter does not yield significant improvements, except in the \texttt{Waterway Long} sequence. 

Surprisingly, all LiDAR–Inertial approaches based on Kalman filtering fail to outperform LiDAR-only methods (e.g.,\ \cite{genz-icp}) on the \texttt{Uneven Tunnel5} sequence. This outcome is likely due to high IMU noise caused by the uneven, under-construction tunnel floor. The \texttt{Bridge1} sequence demonstrates the most challenging scenario, where all benchmark methods fail due to its featureless, repetitive structure combined with high-speed motion. Although the accuracy of our proposed methods is not satisfactory in this case, they maintain stability and do not diverge, unlike the baselines, even in such adverse conditions.

\textbf{\textit{Discussion:}} Based on the experimental observations above, we have successfully demonstrated effective uncertainty estimation of the proposed \SVNICP, as exemplified by the challenging corridor scenario in Fig.\,\ref{fig: long_corridor}. In this environment, the estimated ICP uncertainties along each dimension consistently align with the environmental structure and the robot’s motion, enabling realistic Kalman gain scaling in a simple Kalman filter when the LiDAR degrades. Besides, Fig.\,\ref{fig: long_corridor} b) and c) show that the error samples for each state dimension remain bounded within the $3\sigma$ confidence intervals. As a result, both APE and overall robustness are significantly improved without the need for hand-crafted noise models, which aligns with our expectation of uncertainty-aware sensor fusion.

In the LIO setup, IMU noise does not influence \SVNICPn’s uncertainty estimation but directly affects the Kalman gain. Thus, setting (roughly) realistic IMU noise values reflecting the robot’s motion is crucial for odometry performance. 

However, while the proposed \SVNICP provides consistent one-shot uncertainty estimation under common LiDAR degradations—such as poor initialization or geometry—it is still limited in certain scenarios when the robot's motion becomes unobservable. These include unstructured environments with aggressive motion (e.g., Laurel Caverns) and LiDAR-degraded scenes with moving objects (e.g., Urban Tunnel), leading to severe map corruption. In such cases, although the estimated uncertainty using the proposed method increases consequentially, incorporating additional observations, such as the robot’s velocity, along with sophisticated map management becomes essential for achieving high-quality odometry performance.

\makeKLNNETable

\subsection{Analysis of Uncertainty Estimation}
To intuitively assess uncertainty quality, we adopted the Monte Carlo-based method from \cite{icp-cov-martin}, using 1000 samples to approximate a pseudo-true ICP distribution\footnote{[Online] https://github.com/CAOR-MINES-ParisTech/3d-icp-cov.}. On the \texttt{Long Corridor} sequence \cite{Data_subtmrs}, we compare our proposed uncertainty estimates against classic methods\footnote{Both the MC-based and classical ICP methods are initialized with the ground-truth transformation, following \cite{icp-cov-martin}. In contrast, our ICP variants start from inaccurate initial state predictions. The pseudo–true covariance estimated with 1000 MC samples provides only an approximate reference for comparison.} \cite{censi, icp-cov-martin} by computing the Kullback-Leibler (KL) divergence and Normalized Norm Error (NNE) \cite{icp-cov-martin}. While the KL divergence evaluates the approximated ICP posterior distribution, the NNE measures the estimation error normalized by the predicted uncertainty, offering additional insight into the accuracy and consistency of both the ICP estimates and their associated uncertainties.

Results are shown in Table \ref{tab:kl_nne}. 
It can be observed that our ICP variants outperform all classical methods \cite{censi, icp-cov-martin} on both KL divergence and NNE, indicating that our uncertainty estimates are accurate and consistent throughout the entire test sequence. Compared to \SVGDICP, which exhibits slow and unstable convergence, \SVNICP provides superior uncertainty estimates, achieving lower KL divergence for both translation and rotation components.

Notably, the proposed \SVNICP in the LIO setting achieves the lowest error metrics, benefiting from both the estimated noise parameters and the accurate initial guess provided by the IMU. In contrast, the significantly higher KL divergence and NNE indicate that a Kalman filter with fixed ICP noise parameters fails to deliver robust and accurate pose estimates.

\begin{figure}[!t]
    \centering
    \includegraphics[width=0.48\textwidth]{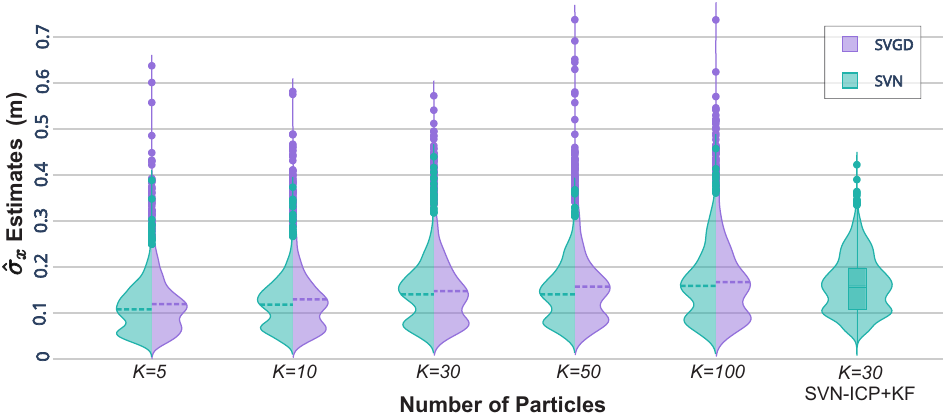}
    \caption{$1\sigma$ estimates in the $x$ direction with varying particles $K$. }
    \label{fig: violin_sigma_x}
\end{figure}

In addition, Fig.\,\ref{fig: violin_sigma_x} visualizes the $1\sigma$ estimates in the $x$ direction obtained using both \SVGDICP and \SVNICP with varying particle sizes\footnote{Only the $x$-axis is shown due to space constraints, as it is the most affected dimension in the corridor scenario.}. It can be observed that the $1\sigma$ estimates in the $x$ direction approximate a non-Gaussian noise distribution characterized by multi-modality and long tails, which aligns with the expected LiDAR degradation in a long corridor environment in Fig.\,\ref{fig: long_corridor}. Moreover, \SVGDICP produces larger noise in the $x$ direction due to unstable convergence, while \SVNICP, fused with a Kalman filter, suppresses long-tail effects thanks to accurate initialization using IMU.

\subsection{Ablation Study on \SVNICP}

\subsubsection{Convergence Analysis of SVGD and SVN} \label{sec: ab_svgd_svn}
\begin{figure}[!t]
    \centering
    \includegraphics[width=0.48\textwidth]{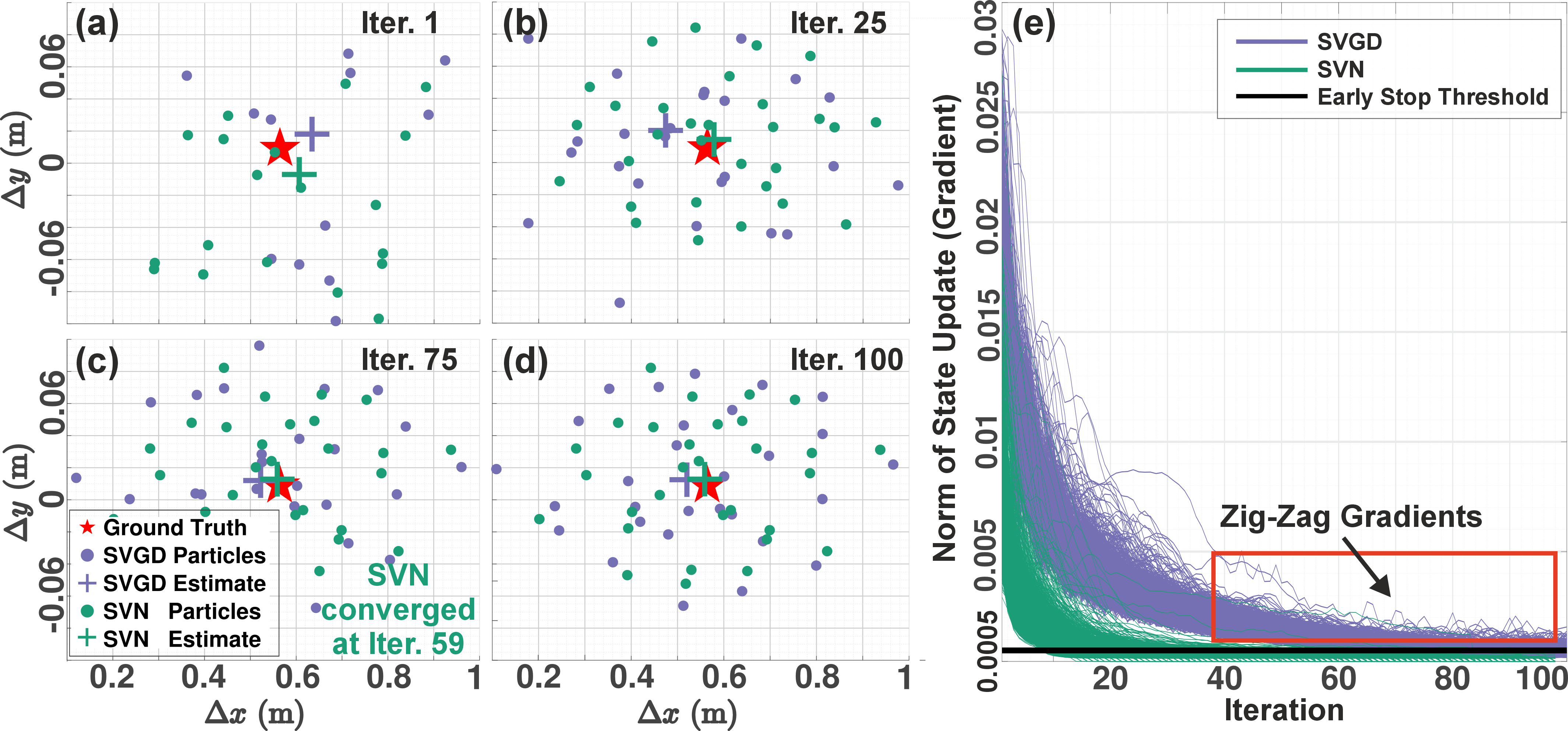}
    \caption{Convergence analysis of \SVGDICP and \SVNICP on the \texttt{Long Corridor} sequence: (a-d) particle distributions of frame 1136, (e) norms of state updates over successive iterations. While \SVNICP\ converges around the 75th iteration, the particles of \SVGDICP\ continue to exhibit noticeable movement at 100th iteration.}
    \label{fig: covergence}
\end{figure}

Fig.\,\ref{fig: covergence} shows SVGD and SVN convergence on the \texttt{Long Corridor} sequence via particle distributions and state-update norms.

Compared to SVN, which converges rapidly within the first 30 iterations (Fig.\,\ref{fig: covergence}b), \SVGDICP shows slower and sometimes ill-conditioned gradient descent, leading to zigzag particle trajectories. As illustrated in Fig.\,\ref{fig: covergence}a, all \SVNICP particles stabilize by the 59th iteration, whereas those of \SVGDICP continue moving until the maximum iteration. This demonstrates the limitation of gradient-based optimization under challenging conditions. Consequently, \SVGDICP generally requires more iterations and runtime (Table~\ref{tab:ablation_particle}) and often fails to meet the early-stopping criterion.

\subsubsection{Particle Size and Runtime}
Table~\ref{tab:ablation_particle} compares \SVGDICP and \SVNICP with different particle counts to assess the trade-off between efficiency and uncertainty accuracy.

As the number of particles increases, both SVGD and SVN achieve lower KL divergence and NNE in solving ICP, reflecting improved distribution approximation—albeit with higher computational cost, as expected for non-parametric, sampling-based methods. However, this improvement does not translate to better ICP accuracy, which remains comparable for \SVGDICP and \SVNICP across all particle counts. Overall, \SVNICP demonstrates superior performance in terms of computation time and convergence speed.

We used 30 particles in all experiments across different test sequences. However, as shown in Fig.\,\ref{fig: violin_sigma_x}, the shape of the approximated distributions remains largely invariant to the number of particles. This suggests that using just 5 to 10 particles is sufficient for robust sensor fusion, while preserving computational efficiency for online applications.
 
\makeAblationTable

%% file: tables/make_subtmrs_table.tex
\newcommand{\makeSubTMRSTable}{

\setlength{\tabcolsep}{3pt}

\begin{table*}[t]
\centering
\caption{General performance metrics on the \texttt{SubT-MRS} dataset \cite{Data_subtmrs}, with accuracy reported in APE~/~RPE (m) format. \textbf{RunTime} indicates the average processing time per LiDAR scan in seconds.}
\scalebox{0.67}{
\begin{threeparttable}
\begin{tabular}{@{}l|l|ccccc|ccc|ccc|cc@{}}
\toprule

\multicolumn{2}{c|}{\textbf{Scene / Robot Type}} &\multicolumn{5}{c|}{ \textbf{Geometric Degradation (UGV)}} &\multicolumn{3}{c|}{ \textbf{Simulation (Drone)}} &\multicolumn{3}{c|}{ \textbf{Mix Degradation}}&&\\

\cmidrule(r{0.55em}){1-2}  \cmidrule(lr{0.55em}){3-7} \cmidrule(lr{0.55em}){8-10} \cmidrule(lr{0.55em}){11-13} \cmidrule(lr{0.55em}){14-15} 
 
\multicolumn{2}{c|}{\textbf{Method}} &  \begin{tabular}[c]{@{}c@{}}Urban \\ ($\SI{1680.2}{s}$)\end{tabular}  &  \begin{tabular}[c]{@{}c@{}}Tunnel \\ ($\SI{3425.4}{s}$)\end{tabular} &  \begin{tabular}[c]{@{}c@{}}Cave \\ ($\SI{1786.1}{s}$)\end{tabular} &  \begin{tabular}[c]{@{}c@{}}Nuclear1 \\ ($\SI{513.0}{s}$)\end{tabular}  &  \begin{tabular}[c]{@{}c@{}}Nuclear2 \\ ($\SI{3177.0}{s}$)\end{tabular} &  \begin{tabular}[c]{@{}c@{}}Factory \\ ($\SI{160.6}{s}$)\end{tabular} & \begin{tabular}[c]{@{}c@{}}Ocean \\ ($\SI{127.4}{s}$)\end{tabular} & \begin{tabular}[c]{@{}c@{}}Sewerage \\ ($\SI{130.9}{s}$)\end{tabular} & \begin{tabular}[c]{@{}c@{}}Laurel Caverns (H) \\ ($\SI{955.3}{s}$)\end{tabular}  & \begin{tabular}[c]{@{}c@{}}Long Corridor (R) \\ ($\SI{280.1}{s}$)\end{tabular} & \begin{tabular}[c]{@{}c@{}}Multi Floor (L) \\ ($\SI{417.6}{s}$)\end{tabular} & \textbf{Average} 
& \textbf{RunTime}\\ 

\cmidrule(r{0.55em}){1-2}  \cmidrule(lr{0.55em}){3-3} \cmidrule(lr{0.55em}){4-4} \cmidrule(lr{0.55em}){5-5} \cmidrule(lr{0.55em}){6-6} \cmidrule(lr{0.55em}){7-7} \cmidrule(lr{0.55em}){8-8} \cmidrule(lr{0.55em}){9-9} \cmidrule(lr{0.55em}){10-10} \cmidrule(l{0.55em}){11-11} \cmidrule(lr{0.55em}){12-12} \cmidrule(lr{0.55em}){13-13}  \cmidrule(lr{0.55em}){14-14} \cmidrule(lr{0.55em}){15-15} 

\multirow{4}{*}{\begin{sideways}system\end{sideways}}
&Liu et al. \cite{FastLio2}\textsuperscript{*} 
& 0.307~/~0.038  & 0.095~/~0.032 & 0.629~/~0.055 & 0.122~/~0.028 & 0.235~/~0.048    &    {\textbf{0.889}}~/~0.191 & 0.757~/~0.174 & {\textbf{0.978}}~/~0.188  & 1.454~/~0.088 & 1.454~/~0.088 & {\textbf{0.401}}~/~0.059 &  {\textbf{0.588}}~/~0.091 
&  51.31\\

&Weitong et al. \cite{FastLio}\textsuperscript{*} 
& {\textbf{0.260}}~/~0.038 & 0.096~/~0.032         & {0.617}~/~0.056           & {0.120}~/~0.029       &  {\textbf{0.222}}~/~0.049       & 0.998~/~0.190  & 0.770~/~0.183    & 1.586~/~0.243  & 0.402~/~\textbf{0.046}       & {1.254}~/~0.086 & 0.577~/~\textbf{0.054} & 0.663~/~0.097 
& 0.125\\

&Kim et al \cite{FastLio2}\textsuperscript{*}  
& 0.331~/~0.098          & {0.092}~/~0.032          & 0.787~/~0.055      & 0.123~/~0.028        & 0.270~/~0.054          & 10.63~/~0.861          & 22.43~/~0.535        & 7.147~/~0.401        & \textbf{0.279}~/~\textbf{0.046}    & 2.100~/~0.093   & 0.650~/~0.260 & 3.825~/~0.219 
& 0.268\\

&Zhong et al. \cite{DLO}\textsuperscript{*} 
& 1.205~/~0.157 & 0.695~/~0.062 & -~/~- & 1.175~/~0.079 & 1.720~/~0.106 & {\textbf{0.889}}~/~0.706 & 0.778~/~0.691 & 1.130~/~0.617 & 2.080~/~0.094 & -~/~-  & -~/~- & -~/~- 
& 0.027\\

\cmidrule(r{0.55em}){1-1} \cmidrule(lr{0.55em}){2-2} \cmidrule(lr{0.55em}){3-3} \cmidrule(lr{0.55em}){4-4} \cmidrule(lr{0.55em}){5-5} \cmidrule(lr{0.55em}){6-6} \cmidrule(lr{0.55em}){7-7} \cmidrule(lr{0.55em}){8-8} \cmidrule(lr{0.55em}){9-9} \cmidrule(lr{0.55em}){10-10} \cmidrule(l{0.55em}){11-11} \cmidrule(lr{0.55em}){12-12} \cmidrule(lr{0.55em}){13-13}  \cmidrule(lr{0.55em}){14-14} \cmidrule(lr{0.55em}){15-15} 

\multirow{4}{*}{\begin{sideways}odom\end{sideways}}&LIO-EKF \cite{LIO-EKF}\textsuperscript{*}  
& 1.060~/~0.130  &0.220~/~0.090  & 0.750~/~0.150  & 0.470~/~0.130   & 0.620~/~0.200         & 4.920~/~\color{UnderWaterDark}{\textbf{0.040}} & {0.280}~/~0.040   & 24.46~/~0.160  & 9.140~/~0.200  & 2.990~/~0.630 & 5.500~/~0.280 & 4.312~/~0.186 
& 0.006\\

&KISS-ICP \cite{kiss-icp} 
& 13.16~/~0.108 & 7.739~/~0.111 & 26.90~/~0.249 & 0.143~/~0.044 & 31.13~/~0.241 & -~/~- & -~/~- & -~/~- & 31.57~/~0.215 & 6.204~/~0.255  & 15.41~/~0.214 & -~/~- 
& 0.005\\

&GenZ-ICP \cite{genz-icp} 
& 0.600~/~0.023 & 0.101~/~0.021 & 0.564~/~0.037 & 0.555~/~0.033 & -~/~- & -~/~- & -~/~- & -~/~- & 0.399~/~\color{UnderWaterDark}{0.050} & 1.746~/~0.073 & 20.49~/~0.101 & -~/~- 
&0.042 \\

&\steinICP \cite{stein-icp} 
& 23.26~/~0.057 &  -~/~- &  -~/~- & -~/~- & -~/~- & -~/~- & -~/~- & -~/~- & -~/~- & -~/~- & -~/~- & -~/~- 
& 0.935 \\

\cmidrule(r{0.75em}){1-1} \cmidrule(lr{0.55em}){2-2} \cmidrule(lr{0.55em}){3-3} \cmidrule(lr{0.55em}){4-4} \cmidrule(lr{0.55em}){5-5} \cmidrule(lr{0.55em}){6-6} \cmidrule(lr{0.55em}){7-7} \cmidrule(lr{0.55em}){8-8} \cmidrule(lr{0.55em}){9-9} \cmidrule(lr{0.55em}){10-10} \cmidrule(l{0.55em}){11-11} \cmidrule(lr{0.55em}){12-12} \cmidrule(lr{0.55em}){13-13}  \cmidrule(lr{0.55em}){14-14} \cmidrule(lr{0.55em}){15-15} 

\multirow{4}{*}{\begin{sideways}odom, ours\end{sideways}} 
&\SVGDICP
& 0.562~/~\color{UnderWaterDark}{\textbf{0.022}} & {\color{UnderWaterDark}{\textbf{0.077}}}~/~{\color{UnderWaterDark}{\textbf{0.018}}} & 0.764~/~\color{UnderWaterDark}{\textbf{0.033}} & 0.141~/~0.024 & -~/~- & -~/~- & 22.57~/~1.050& 20.88~/~0.570 & 16.79~/~0.067 & 1.334~/~0.080& 14.89~/~0.092 & -~/~- 
& 0.669 \\

&\SVNICP 
& 0.551~/~0.031 & 0.110~/~0.028 & 0.562~/~0.042 & 0.131~/~0.026 & -~/~- & -~/~- & 21.97~/~0.869 & 20.42~/~0.575  & {8.911}~/~0.085 & 1.321~/~\color{UnderWaterDark}{\textbf{0.076}}  & 11.15~/~0.088 & -~/~- 
&0.483 \\

&\SVNICPn+KF-fix
& {\color{red}{2.365}}~/~0.025 & 0.081~/~0.028 & {\color{red}{5.260}}~/~0.037 & \color{UnderWaterDark}{\textbf{0.058}}~/~\color{UnderWaterDark}{\textbf{0.017}} & 0.464~/~\color{UnderWaterDark}{\textbf{0.027}} & {\color{UnderWaterDark}{4.570}}~/~0.137 & {\color{red}1.173}~/~0.033 & {\color{red}11.51}~/~0.107 & {{5.197}}~/~0.056  & {\color{red}1.567}~/~0.086 & {\color{UnderWaterDark}{0.692}}~/~0.054 & {\color{red}2.995}~/~0.055 & 0.568 \\

&\SVNICPn+KF 
& {\color{UnderWaterDark}{0.478}}~/~0.030 & 0.085~/~0.027  & {\color{UnderWaterDark}{\textbf{0.532}}}~/~0.040 & 0.059~/~0.023 & {\color{UnderWaterDark}{0.441}}~/~0.046 & 4.836~/~0.053 & \color{UnderWaterDark}{\textbf{0.047}}~/~\color{UnderWaterDark}{\textbf{0.031}} & \color{UnderWaterDark}{2.933}~/~\color{UnderWaterDark}{\textbf{0.068}} & {\color{UnderWaterDark}{4.089}}~/~{0.056} & {\color{UnderWaterDark}{\textbf{0.650}}}~/~0.089  & 0.911~/~{\color{UnderWaterDark}{0.053}}  & \color{UnderWaterDark}{1.369}~/~0.046 &   0.515\\

\bottomrule
\end{tabular}
\begin{tablenotes}
      \small
      \item[1] Best results are shown in \textbf{bold}. The best among odometry-only methods are highlighted in {\color{UnderWaterDark}blue}, while {\color{red}red} indicates notable differences when using fixed noise parameters (SVN-ICP+KF-fix).
      \item[2] Results indicated by ``*" are reported as presented in \cite{Data_subtmrs}.
      \item[3] Methods labeled as ``system" are cascaded pipelines with multiple localization approaches, often fine-tuned with loop-closure constraints for competition purposes \cite{Data_subtmrs}. 
      \item[4] Methods labeled as ``odom" refer to pure LiDAR or LiDAR–Inertial odometry without loop closure and sophisticated engineering.
      \item[5] \SVGDICP adopts the \SVNICP pipeline but replaces the descent direction with SVGD instead of second-order SVN.
      \item[6] H, R, and L denote hand-held, RC car, and legged robot platforms, respectively. We excluded the Block LiDAR sequence from our experiments due to the absence of ground-truth data.
      \item[7] We omitted results where the algorithm failed across multiple test runs or produced a large APE (e.g., $>\SI{100}{m}$); these cases are denoted as ``–".
      \item[8] By default, \steinICP, \SVGDICP, and \SVNICP use 30 particles optimized over a maximum of 100 iterations. All other parameters are kept identical to ensure a fair benchmark across all experiments.
    \end{tablenotes}
\end{threeparttable}
}
\label{tab:ATE_RPE_SUBTMRS}
\end{table*}

}

%% file: tables/make_geode_table.tex
\newcommand{\makeGEODETable}{

\setlength{\tabcolsep}{3pt}
\begin{table}[t]
\centering
\caption{APE (m) on the \texttt{GEODE} dataset \cite{Data_geode}.}
\scalebox{0.81}{
\begin{threeparttable}
\begin{tabular}{@{}c|l|cc|cc|c|c@{}}
\toprule

\multicolumn{2}{c|}{\textbf{Robot Type}} &\multicolumn{2}{c|}{\textbf{UGV}} &\multicolumn{2}{c|}{\textbf{Sailboat}} &\multicolumn{1}{c|}{\textbf{Handheld}}&\multicolumn{1}{c}{\textbf{Vehicle}}\\

\cmidrule(r{0.55em}){1-2}  \cmidrule(lr{0.55em}){3-4} \cmidrule(lr{0.55em}){5-6} \cmidrule(lr{0.55em}){7-7} \cmidrule(lr{0.55em}){8-8} 
\multicolumn{2}{c|}{\multirow{2}{*}{\textbf{Method}}} &  Offroad1 &  \begin{tabular}[c]{@{}c@{}}Uneven \\ Tunnel5 \end{tabular} & \begin{tabular}[c]{@{}c@{}}Waterway \\ Short\end{tabular} &  \begin{tabular}[c]{@{}c@{}}Waterway \\ Long\end{tabular}  & Stair  & Bridge1 \\ 

\multicolumn{2}{c|}{} &  ($\SI{426.6}{s}$) & ($\SI{244.3}{s}$) & ($\SI{472.0}{s}$) & ($\SI{1616.7}{s}$) & ($\SI{345.5}{s}$)  & ($\SI{382.5}{s}$) \\ 

\cmidrule(r{0.55em}){1-2}  \cmidrule(lr{0.55em}){3-3} \cmidrule(lr{0.55em}){4-4} \cmidrule(lr{0.55em}){5-5} \cmidrule(lr{0.55em}){6-6} \cmidrule(lr{0.55em}){7-7} \cmidrule(lr{0.55em}){8-8} 

\multirow{4}{*}{\begin{sideways}system\end{sideways}}
& LIO-SAM \cite{liosam}\textsuperscript{*}      & 0.23  & 0.16  &  -    &  -      & 6.30 & - \\
& LVI-SAM \cite{LVI-SAM}\textsuperscript{*}     & -     & 0.30  & -     &  -     & -    & - \\
& R3LIVE \cite{R3LIVE}\textsuperscript{*}       & -     & 11.44 & 4.80  &  -     & 4.54 & - \\
& RELEAD \cite{RELEAD}\textsuperscript{*}       & 0.22  & 0.17  & 8.5   & 67.40   & \textbf{0.57} & - \\

\cmidrule(r{0.55em}){1-2}  \cmidrule(lr{0.55em}){3-3} \cmidrule(lr{0.55em}){4-4} \cmidrule(lr{0.55em}){5-5} \cmidrule(lr{0.55em}){6-6} \cmidrule(lr{0.55em}){7-7} \cmidrule(lr{0.55em}){8-8} 

\multirow{3}{*}{\begin{sideways}odom\end{sideways}}
& FAST-LIO2 \cite{FastLio2}\textsuperscript{*}  & 0.18  & 0.19  & 10.27 & 70.26 & 4.69 & - \\
& DLIO \cite{DLIO}\textsuperscript{*}           & 0.23  & 0.13  & 2.59  & 68.04  & 4.89 & - \\
& GenZ-ICP \cite{genz-icp}  & -     & \textbf{0.11}    &  -    &  -     &   -  &  - \\

\cmidrule(r{0.55em}){1-2}  \cmidrule(lr{0.55em}){3-3} \cmidrule(lr{0.55em}){4-4} \cmidrule(lr{0.55em}){5-5} \cmidrule(lr{0.55em}){6-6} \cmidrule(lr{0.55em}){7-7} \cmidrule(lr{0.55em}){8-8} 

\multirow{4}{*}{\begin{sideways}odom, ours\end{sideways}}
& \SVGDICP               & 7.06          & 0.14  & 1.80          & 39.28          & 10.1 &   -    \\
& \SVNICP                & 0.25          & 0.15  & 2.97          & 29.86          & 7.90 & 92.23  \\
& \SVNICPn+KF-fix        & 0.13          & 0.22  & 1.82          & 36.80          & 2.89 & 45.08  \\
& \SVNICPn+KF            & \textbf{0.12} & 0.21  & \textbf{1.28} & \textbf{15.58} & 2.80 &  \textbf{44.27}  \\

\bottomrule
\end{tabular}
\begin{tablenotes}
      \smaller
      \item[1] Best results are shown in \textbf{bold}. 
      \item[2] Results indicated by ``*" are reported as presented in \cite{Data_geode}.
      \item[3] We omitted results where the algorithm failed or produced a APE larger than 100m; denoted as ``–".
      \item[4] We selected only a subset of test sequences with the \texttt{alpha} sensor setting, based on Table 6 in \cite{Data_geode}.
      \item[5] In the absence of additional measurements, all algorithms fail in urban tunnel and bridge sequences.
    \end{tablenotes}
\end{threeparttable}
}
\label{tab:ATE_RPE_GEODE}
\end{table}
}

%% file: tables/make_ablation_table.tex
\newcommand{\makeAblationTable}{
\begin{table}[t]
  \centering
  \caption{Ablation study on number of particles $K$ for both SVGD- and SVN-ICP on the \texttt{Long Corridor} sequence.}
  \label{tab:ablation_particle}
  \scalebox{.85}{
\begin{threeparttable}
 \begin{tabular}{c|l|cccccc}
    \toprule
    \multirow{3}{*}{} & \multicolumn{1}{c}{\multirow{3}{*}{\textbf{Metric}}} & \multicolumn{6}{c}{\textbf{Number of Particles}} \\
    \cmidrule(lr){3-8}
    & & \textbf{1} & \textbf{5} & \textbf{10} & \textbf{30} & \textbf{50} & \textbf{100}  \\
    
   \cmidrule(r{0.55em}){1-2}  \cmidrule(lr{0.55em}){3-8} 
    
    \multirow{6}{*}{\begin{sideways}~\SVGDICP\end{sideways}}
        & APE (m)     & 1.165          & 1.267 & 1.484 & 1.324 & 1.378          & 1.349 \\
        & Runtime (s) &  0.486         & 0.500 & 0.667 & 0.611 &  0.771         & 1.209 \\
        & Avg. iter.  &  96            &  98   &  99   & 99    & 99            &  96  \\
        \cmidrule(lr){2-2} \cmidrule(lr){3-8}
        & KL div.     &  - / -            & 5.00 / 928.5 & 1.59 / 209 & 1.58 / 79.1 & 1.62 / 48.3          & 1.66 / 7.28 \\
        & NNE &  - / -         & 1.30 / 1.46      & 1.19 / 1.03      & 0.99 / 0.64      &   0.93 / 0.58             & 0.88 / 0.53      \\
        
    \cmidrule(r{0.55em}){1-2}  \cmidrule(lr{0.55em}){3-8} 
    
    \multirow{6}{*}{\begin{sideways}~\SVNICP\end{sideways}}
    & APE (m)     & 1.205             & 1.256 & 1.270 & 1.321 & 1.268 & 1.383 \\
    & Runtime (s) & 0.251             & 0.337  & 0.369 & 0.469 & 0.519         & 0.691 \\
    & Avg. iter.  &   33              &   56   &   65  &  65  &  64           &  62   \\
    \cmidrule(lr){2-2} \cmidrule(lr){3-8}
    & KL div.     &   - / -          & 3.94 / 1073      &  1.58 / 207     & 1.53 / 41.7      & 1.58 / 48.3 & 1.60 / 13.7      \\
    & NNE  &  - / -                 & 1.37 / 1.76      &  1.21 / 1.22     & 0.99 / 0.85      & 0.97 / 0.77 & 0.88 / 0.70      \\
   
    \bottomrule
  \end{tabular}
  \begin{tablenotes}
      \smaller
      \item[1] KL div. and NNE computed with the 90th MC quantil cut-off (see Table \ref{tab:kl_nne}). 
      \item[2] KL div. and NNE are reported in translation / rotation format.  
      \item[3] Runtime measures the computation time of Algorithm I, excluding all pre- and postprocessing steps.
    \end{tablenotes}
\end{threeparttable}
  }
\end{table}
}

%% file: tables/make_KL_NNE_table.tex
\newcommand{\makeKLNNETable}{
\begin{table}[t]
\centering
\caption{Normalized Norm Error (NNE) and Kullback–Leibler divergence (KL div.) for translation and rotation against 1000 Monte Carlo ICP samples on the \texttt{Long Corridor} sequence \cite{Data_subtmrs}.
}
\label{tab:kl_nne}
\scalebox{.85}{
\begin{threeparttable}
\begin{tabular}{l|c|c|c|c|c|c|c|c}
\toprule

\multirow{2}{*}{~~~\textbf{Uncertainty}} &  \multicolumn{2}{c|}{\textbf{NNE}}  & \multicolumn{2}{c|}{\textbf{KL div.}} &  \multicolumn{2}{c|}{\textbf{NNE*}}  & \multicolumn{2}{c}{\textbf{KL div.*}}   \\

&   ~trans.~ &   ~~rot.~~ &  ~trans.~ & ~~rot.~~ &  ~trans.~ &  ~~rot.~~  & ~trans.~  & ~~rot.~~\\
\cmidrule(r{0.55em}){1-1}  \cmidrule(lr{0.55em}){2-3} \cmidrule(lr{0.55em}){4-5} \cmidrule(lr{0.55em}){6-7} \cmidrule(lr{0.55em}){8-9} 
\vspace{.4em}
$\myFrameVecHat{\Sigma}{\text{Censi}}{}$ \cite{censi}          &  126.5             & 565.4 &   1.2e5   &  5.8e5     &  101.2  & 336.6 & 1.0e5  &  2.8e5 \\
 
\vspace{.4em}
$\myFrameVecHat{\Sigma}{\text{COV-3D}}{}$ \cite{icp-cov-martin}&  1.486             & 2.292  &   62.21    &  6.3e3    &  1.088  &  1.644 & 12.68 & 385.8 \\

\cmidrule(r{0.55em}){1-1}  \cmidrule(lr{0.55em}){2-3} \cmidrule(lr{0.55em}){4-5} \cmidrule(lr{0.55em}){6-7} \cmidrule(lr{0.55em}){8-9} 

\vspace{.4em}
$\myFrameVecHat{\Sigma}{\text{\SVGDICP}}{}$                    &  1.288          & \textbf{1.284} &   2.067   &   272.9    &  0.997  &  \textbf{0.637}& 1.580&   79.10\\

\vspace{.4em}
$\myFrameVecHat{\Sigma}{\text{\SVNICP}}{}$                     &  1.281          & 1.855 &   \textbf{1.996}   &  93.36    &  0.995  & 0.852 &  \textbf{1.532}&  41.72\\

\cmidrule(r{0.55em}){1-1}  \cmidrule(lr{0.55em}){2-3} \cmidrule(lr{0.55em}){4-5} \cmidrule(lr{0.55em}){6-7} \cmidrule(lr{0.55em}){8-9} 

\vspace{.4em}
$\myFrameVecHat{\Sigma}{\text{\SVNICP-KF-fix}}{}$              &  {\color{red}27.56}          & {\color{red}19.02} &   {\color{red}1.2e3}   &  {\color{red}1.1e3}    &  {\color{red}15.90}  & {\color{red}9.004} & {\color{red}590.7} &  {\color{red}465.0} \\

\vspace{.4em}
$\myFrameVecHat{\Sigma}{\text{\SVNICP-KF}}{}$                  &  \textbf{1.196} &  1.559 &   2.181   &  \textbf{25.37}    &  \textbf{0.939}  & 0.675 & 1.713& \textbf{10.06}\\
\bottomrule
\end{tabular}
\begin{tablenotes}
      \smaller
      \item[1] ``*" denotes results with a 90th quantil cut-off, removing unstable MC estimates.
      \item[2] Best results are shown in \textbf{bold}. 
      \item[3] {\color{red}red} indicates notable differences when fusing SVN-ICP with fixed noise parameters. 
    \end{tablenotes}
\end{threeparttable}}
\end{table}
}

%% file: sec/7_conclusion_acknowledgement.tex
\section{Conclusion}
This letter presents \SVNICP, a novel ICP-based LiDAR odometry method leveraging Stein Variational Newton, enabling accurate transformation and consistent uncertainty estimation. Evaluated on challenging datasets against state-of-the-art methods, the proposed approach achieves superior accuracy and robustness, highlighting that realistic uncertainty estimation is essential for robust sensor fusion.

However, the current implementation of \SVNICP employs a basic odometry design with a simple Kalman filter and naïve map management, pointing to future research topics toward highly efficient, uncertainty-aware LiDAR odometry that supports active SLAM. In addition, ongoing work also includes investigating the kernel parameterization and extending the \SVNICP to other range sensors, such as radar.

\clearpage
\section*{ACKNOWLEDGMENT}
Special thanks to Shibo Zhao and Honghao Zhu from the AirLab at Carnegie Mellon University for their support with the \texttt{SubT-MRS} dataset and insightful discussions on the method. We also thank Fahira Afzal Maken and Prof. Fabio Ramos from the School of Computer Science at the University of Sydney for their valuable input during the early stages of this work.